\documentclass[letterpaper]{article} 
\usepackage{aaai23}  
\usepackage{times}  
\usepackage{helvet}  
\usepackage{courier}  
\usepackage[hyphens]{url}  
\usepackage{graphicx} 
\usepackage{natbib}  
\usepackage{caption} 
\usepackage{algorithm}
\usepackage{algorithmic}

\usepackage{newfloat}
\usepackage{listings}
\DeclareCaptionStyle{ruled}{labelfont=normalfont,labelsep=colon,strut=off} 
\floatstyle{ruled}
\newfloat{listing}{tb}{lst}{}
\floatname{listing}{Listing}
\newcommand{\repeatthanks}{\textsuperscript{\thefootnote}}

\title{KnowGL: Knowledge Generation and Linking from Text}
\author{
   Gaetano Rossiello\thanks{Equal contributions.}, Md. Faisal Mahbub Chowdhury\repeatthanks, Nandana Mihindukulasooriya,\\Owen Cornec, Alfio Massimiliano Gliozzo
}
\affiliations{
    IBM Research AI\\Thomas J. Watson Research Center, NY
}

\begin{document}

\maketitle

\begin{abstract}
We propose KnowGL, a tool that allows converting text into structured relational data represented as a set of ABox assertions compliant with the TBox of a given Knowledge Graph (KG), such as Wikidata. 
We address this problem as a sequence generation task by leveraging pre-trained sequence-to-sequence language models, e.g. BART.
Given a sentence, we fine-tune such models to detect pairs of entity mentions and jointly generate a set of facts consisting of the full set of semantic annotations for a KG, such as entity labels, entity types, and their relationships.
To showcase the capabilities of our tool, we build a web application consisting of a set of UI widgets that help users to navigate through the semantic data extracted from a given input text. We make the KnowGL model available at~\url{https://huggingface.co/ibm/knowgl-large}.
\end{abstract}

\section{Introduction and Related Work}
A Knowledge Graph (KG) is defined as a semantic network where entities, such as objects, events or concepts, are connected between them through relationships or properties.
KGs are organized in multi-graph data structures and stored as a set of triples (or facts), i.e. \textsc{(Subject, Relation, Object)}, grounded with a given well-defined ontology~\cite{DBLP:series/synthesis/2021Hogan}.
The usage of formal languages to represent KGs enables unambiguous access to data and facilitates automatic reasoning capabilities that enhance downstream applications, such as analytics, knowledge discovery or recommendations~\cite{DBLP:journals/corr/abs-2207-05188}.

\begin{figure}[t!]
    \centering
    \includegraphics[width=0.34\textwidth]{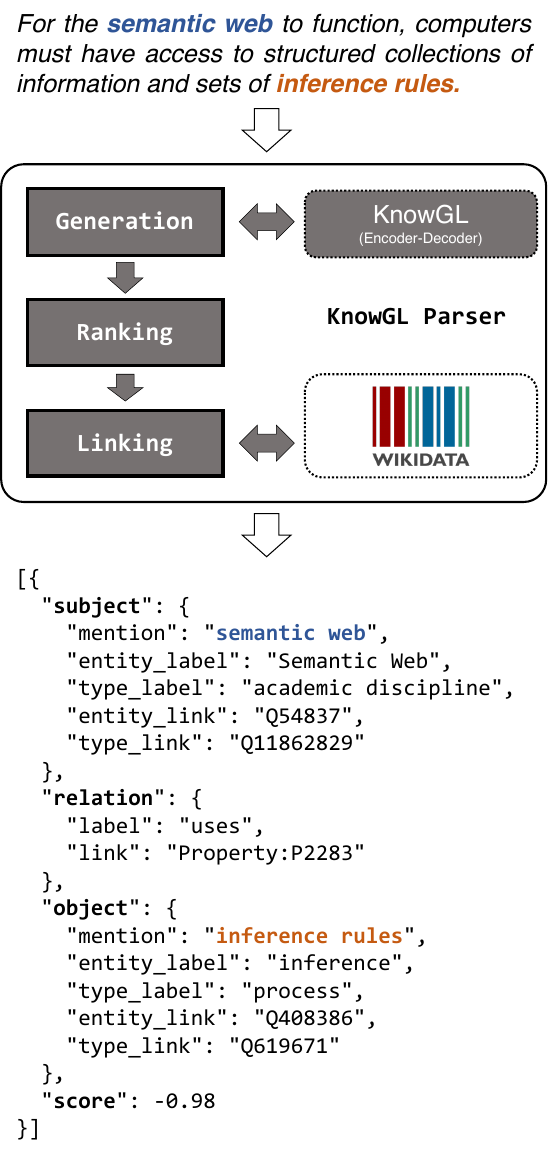}
    \caption{KnowGL Parser Framework}
    \label{fig:knowgl-parser-framework}
\end{figure}

However, building and curating KGs, such as Wikidata~\cite{10.1145/2629489}, requires a considerable human effort.
Systems such as, NELL~\cite{DBLP:conf/aaai/CarlsonBKSHM10}, DeepDive~\cite{DBLP:conf/vlds/NiuZRS12}, Knowledge Vault~\cite{DBLP:conf/kdd/0001GHHLMSSZ14}, DiffBot~\cite{DBLP:conf/emnlp/MesquitaCSMB19} implement Information Extraction (IE) methods for automatic knowledge base population.
A standard IE pipeline consists of several steps, such as co-reference resolution~\cite{DBLP:conf/emnlp/Dobrovolskii21}, named entity recognition~\cite{DBLP:conf/acl/WangJBWHHT20}, relation extraction~\cite{DBLP:conf/naacl/ZhongC21}, and entity liking~\cite{DBLP:conf/emnlp/WuPJRZ20}, each of which is commonly addressed as a separate task.
A pipeline approach presents several limitations, e.g. error propagation among different IE components and complex deployment procedures.
Moreover, each component of the pipeline is trained independently using different architectures and training sets.

The ability of generating structured data from text makes sequence-to-sequence Pre-trained Language Models (PLMs), such as BART~\cite{DBLP:conf/acl/LewisLGGMLSZ20} or T5~\cite{DBLP:journals/jmlr/RaffelSRLNMZLL20},  a valuable alternative to successfully address IE~\cite{DBLP:conf/emnlp/GlassRCG21,DBLP:conf/naacl/GlassRCNCG22,DBLP:journals/corr/abs-2202-13229,DBLP:conf/acl/0001ZL22,DBLP:conf/emnlp/CabotN21,DBLP:conf/naacl/JosifoskiCPP022}, entity/relation linking~\cite{DBLP:conf/iclr/CaoI0P21,DBLP:conf/semweb/RossielloMABGNK21} and semantic parsing~\cite{DBLP:conf/emnlp/ZhouNALFR21,DBLP:conf/www/RongaliSMH20,DBLP:conf/emnlp/DogninPMD21} tasks.  
In this work, we further explore this direction by asking whether PLMs can be fine-tuned to read a sentence and generate the corresponding full set of semantic annotations that are compliant with the terminology of a KG.
For this purpose, we propose a framework able to convert text into a set of Wikidata statements. 
As shown in~\cite{DBLP:journals/corr/abs-2207-05188}, KnowGL parser can be used to automatically extract KGs from collections of documents, with the purpose to help users with semantic content exploration, to create trend analysis, to extract entity infoboxes from text, or to enhance content-based recommendation systems with semantic features. 

\section{KnowGL Parser}
Figure~\ref{fig:knowgl-parser-framework} shows an overview of the Knowledge Generation and Linking (KnowGL) tool. Given a sentence as input, KnowGL returns a list of triples \texttt{(subject, relation, object)} in a JSON format~\footnote{Figure~\ref{fig:knowgl-parser-framework} shows a special case where the output consists of only one triple. However, KnowGL is able to identify multiple pairs of mentions in the input sentence and for each of them, it generates the corresponding triple with the semantic annotations in one-pass computation.}.
From the example, KnowGL identifies the entity mentions \textit{semantic web} and \textit{inference rules} in the sentence and generates the relation label \textit{use} between them. For each mention, the output provides the corresponding entity label and its type. If the entity labels, entity types and relation labels are found in Wikidata, then KnowGL also provides the Wikidata IDs associated with them.
As shown in Figure~\ref{fig:knowgl-parser-framework},
KnowGL Parser consists of three main components: generation, ranking and linking, as described below.

\paragraph{Knowledge Generation}
We address the fact extraction as an autoregressive generation problem. In other words, given a natural language text input, the knowledge generation model generates a linearized sequence representation that contains a set of facts expressed in the textual input.
We adopt the following schema to represent the semantic annotations of a triple in the target sequence: \texttt{[(subject mention \# subject label \# subject type) | relation label | (object mention  \# object label \# object type)]}. If the input text contains multiple mention pairs, the linearized target representations are concatenated using \texttt{\$} as a separator, and the facts are sorted by the order of the appearances of the head entity in the input text.
Unlike in~\cite{DBLP:conf/emnlp/CabotN21,DBLP:conf/naacl/JosifoskiCPP022}, we generate the surface forms, entity labels, and type information for both head and tail entities in the target representation. This represents a full set of semantic annotations, i.e. ABox and TBox, to construct and populate a KG with new facts.
Our hypothesis is such self-contained fact representation also acts as an implicit constraint during decoding.
We exploit BART-large~\cite{DBLP:conf/acl/LewisLGGMLSZ20} as the base model and cast this as a translation task where at training time the encoder receives a sentence, and the decoder generates the sequence target representation as described above. 
To train the generation model, we extend the REBEL dataset~\cite{DBLP:conf/emnlp/CabotN21} by adding the entity labels and their types from Wikidata for each entity surface form in the text.
REBEL is an updated and cleaner version of T-REx~\cite{DBLP:conf/lrec/ElSaharVRGHLS18}, a distantly supervised dataset for relation extraction built by aligning Wikipedia abstracts with Wikidata triples.
We use cross-entropy loss as standard in machine translation whereas, in teacher forcing, the model regards the translation problem as a one-to-one mapping process and maximizes the log-likelihood of generating the linearized facts given the input text.
As reported in~\cite{DBLP:journals/corr/abs-2207-05188}, our KnowGL model outperforms (F1 = 70.74) both a standard IE pipeline system (F1 =  42.50) and the current state-of-the-art generative IE model (F1 = 68.93)~\cite{DBLP:conf/naacl/JosifoskiCPP022}.
For the evaluation, we use the test set released with the REBEL dataset.

\paragraph{Fact Ranking}
This component parses the target sequences generated by the knowledge generation model using a regular expression. The goal is to create a ranked list of distinct facts with their scores. We extract facts from all the returned sequences generated by each beam (where the number of beams is a hyper-parameter). For each extracted fact we consider the negative log-likelihood of the entire generated sequence as a score. Since the same fact can appear in different returned sequences, we sum the scores of each sequence where the fact occurs. The idea is to promote those facts/triples that occur multiple times in different beams. Finally, the facts are sorted by their scores.

\paragraph{Linking to Wikidata}
The linking component enables retrieving the Wikidata IDs associated with the generated entity, type and relation labels. 
For efficiency, we create label-to-IDs maps from Wikidata and store them in key-value data storage systems to avoid bottlenecks caused by running multiple SPARQL queries to a Wikidata triple store. 
It is worth noticing that the model can generate new entity, type, or relation labels that are not in Wikidata. In this case, the linking component returns a \texttt{null} ID and the triple can be used as a candidate for adding new facts in Wikidata.

\section{Demonstration}

KnowGL Parser is implemented in Python language and deployed as a REST API using the Flask framework. The input is a sentence and the output is a JSON format structured as shown in Figure~\ref{fig:knowgl-parser-framework}.
The user interface described in our video demonstration is implemented as a separate web application using Node.js and React web tool frameworks. The UI allows users to insert textual content using a textbox.
Then, the returned JSON is parsed by the UI enabling different types of visualizations. 
For instance, the facts can be organized in a directed multi-graph where the nodes are the entities and the edges represent the relations between two entities. 
The user can navigate and interact with the nodes and edges to easily locate the textual evidence associated with the triples.

\bibliography{refs}

\begin{thebibliography}{25}
\providecommand{\natexlab}[1]{#1}

\bibitem[{Cabot and Navigli(2021)}]{DBLP:conf/emnlp/CabotN21}
Cabot, P.~H.; and Navigli, R. 2021.
\newblock {REBEL:} Relation Extraction By End-to-end Language generation.
\newblock In \emph{{EMNLP} (Findings)}, 2370--2381. Association for
  Computational Linguistics.

\bibitem[{Cao et~al.(2021)Cao, Izacard, Riedel, and
  Petroni}]{DBLP:conf/iclr/CaoI0P21}
Cao, N.~D.; Izacard, G.; Riedel, S.; and Petroni, F. 2021.
\newblock Autoregressive Entity Retrieval.
\newblock In \emph{{ICLR}}. OpenReview.net.

\bibitem[{Carlson et~al.(2010)Carlson, Betteridge, Kisiel, Settles, Jr., and
  Mitchell}]{DBLP:conf/aaai/CarlsonBKSHM10}
Carlson, A.; Betteridge, J.; Kisiel, B.; Settles, B.; Jr., E. R.~H.; and
  Mitchell, T.~M. 2010.
\newblock Toward an Architecture for Never-Ending Language Learning.
\newblock In \emph{{AAAI}}. {AAAI} Press.

\bibitem[{de~S{\'{a}}~Mesquita et~al.(2019)de~S{\'{a}}~Mesquita, Cannaviccio,
  Schmidek, Mirza, and Barbosa}]{DBLP:conf/emnlp/MesquitaCSMB19}
de~S{\'{a}}~Mesquita, F.; Cannaviccio, M.; Schmidek, J.; Mirza, P.; and
  Barbosa, D. 2019.
\newblock KnowledgeNet: {A} Benchmark Dataset for Knowledge Base Population.
\newblock In \emph{{EMNLP/IJCNLP} {(1)}}, 749--758. Association for
  Computational Linguistics.

\bibitem[{Dobrovolskii(2021)}]{DBLP:conf/emnlp/Dobrovolskii21}
Dobrovolskii, V. 2021.
\newblock Word-Level Coreference Resolution.
\newblock In \emph{{EMNLP} {(1)}}, 7670--7675. Association for Computational
  Linguistics.

\bibitem[{Dognin et~al.(2021)Dognin, Padhi, Melnyk, and
  Das}]{DBLP:conf/emnlp/DogninPMD21}
Dognin, P.~L.; Padhi, I.; Melnyk, I.; and Das, P. 2021.
\newblock ReGen: Reinforcement Learning for Text and Knowledge Base Generation
  using Pretrained Language Models.
\newblock In \emph{{EMNLP} {(1)}}, 1084--1099. Association for Computational
  Linguistics.

\bibitem[{Dong et~al.(2014)Dong, Gabrilovich, Heitz, Horn, Lao, Murphy,
  Strohmann, Sun, and Zhang}]{DBLP:conf/kdd/0001GHHLMSSZ14}
Dong, X.; Gabrilovich, E.; Heitz, G.; Horn, W.; Lao, N.; Murphy, K.; Strohmann,
  T.; Sun, S.; and Zhang, W. 2014.
\newblock Knowledge vault: a web-scale approach to probabilistic knowledge
  fusion.
\newblock In \emph{{KDD}}, 601--610. {ACM}.

\bibitem[{ElSahar et~al.(2018)ElSahar, Vougiouklis, Remaci, Gravier, Hare,
  Laforest, and Simperl}]{DBLP:conf/lrec/ElSaharVRGHLS18}
ElSahar, H.; Vougiouklis, P.; Remaci, A.; Gravier, C.; Hare, J.~S.; Laforest,
  F.; and Simperl, E. 2018.
\newblock T-REx: {A} Large Scale Alignment of Natural Language with Knowledge
  Base Triples.
\newblock In \emph{{LREC}}. European Language Resources Association {(ELRA)}.

\bibitem[{Glass et~al.(2021)Glass, Rossiello, Chowdhury, and
  Gliozzo}]{DBLP:conf/emnlp/GlassRCG21}
Glass, M.~R.; Rossiello, G.; Chowdhury, M. F.~M.; and Gliozzo, A. 2021.
\newblock Robust Retrieval Augmented Generation for Zero-shot Slot Filling.
\newblock In \emph{{EMNLP} {(1)}}, 1939--1949. Association for Computational
  Linguistics.

\bibitem[{Glass et~al.(2022)Glass, Rossiello, Chowdhury, Naik, Cai, and
  Gliozzo}]{DBLP:conf/naacl/GlassRCNCG22}
Glass, M.~R.; Rossiello, G.; Chowdhury, M. F.~M.; Naik, A.; Cai, P.; and
  Gliozzo, A. 2022.
\newblock Re2G: Retrieve, Rerank, Generate.
\newblock In \emph{{NAACL-HLT}}, 2701--2715. Association for Computational
  Linguistics.

\bibitem[{Hogan et~al.(2021)Hogan, Blomqvist, Cochez, d'Amato, de~Melo,
  Guti{\'{e}}rrez, Kirrane, Gayo, Navigli, Neumaier, Ngomo, Polleres, Rashid,
  Rula, Schmelzeisen, Sequeda, Staab, and
  Zimmermann}]{DBLP:series/synthesis/2021Hogan}
Hogan, A.; Blomqvist, E.; Cochez, M.; d'Amato, C.; de~Melo, G.;
  Guti{\'{e}}rrez, C.; Kirrane, S.; Gayo, J. E.~L.; Navigli, R.; Neumaier, S.;
  Ngomo, A.~N.; Polleres, A.; Rashid, S.~M.; Rula, A.; Schmelzeisen, L.;
  Sequeda, J.~F.; Staab, S.; and Zimmermann, A. 2021.
\newblock \emph{Knowledge Graphs}, volume~54.

\bibitem[{Josifoski et~al.(2022)Josifoski, Cao, Peyrard, Petroni, and
  West}]{DBLP:conf/naacl/JosifoskiCPP022}
Josifoski, M.; Cao, N.~D.; Peyrard, M.; Petroni, F.; and West, R. 2022.
\newblock GenIE: Generative Information Extraction.
\newblock In \emph{{NAACL-HLT}}, 4626--4643. Association for Computational
  Linguistics.

\bibitem[{Lewis et~al.(2020)Lewis, Liu, Goyal, Ghazvininejad, Mohamed, Levy,
  Stoyanov, and Zettlemoyer}]{DBLP:conf/acl/LewisLGGMLSZ20}
Lewis, M.; Liu, Y.; Goyal, N.; Ghazvininejad, M.; Mohamed, A.; Levy, O.;
  Stoyanov, V.; and Zettlemoyer, L. 2020.
\newblock {BART:} Denoising Sequence-to-Sequence Pre-training for Natural
  Language Generation, Translation, and Comprehension.
\newblock In \emph{{ACL}}, 7871--7880. Association for Computational
  Linguistics.

\bibitem[{Mihindukulasooriya et~al.(2022)Mihindukulasooriya, Sava, Rossiello,
  Chowdhury, Yachbes, Gidh, Duckwitz, Nisar, Santos, and
  Gliozzo}]{DBLP:journals/corr/abs-2207-05188}
Mihindukulasooriya, N.; Sava, M.; Rossiello, G.; Chowdhury, M. F.~M.; Yachbes,
  I.; Gidh, A.; Duckwitz, J.; Nisar, K.; Santos, M.; and Gliozzo, A. 2022.
\newblock Knowledge Graph Induction enabling Recommending and Trend Analysis:
  {A} Corporate Research Community Use Case.
\newblock \emph{CoRR}, abs/2207.05188.

\bibitem[{Ni et~al.(2022)Ni, Rossiello, Gliozzo, and
  Florian}]{DBLP:journals/corr/abs-2202-13229}
Ni, J.; Rossiello, G.; Gliozzo, A.; and Florian, R. 2022.
\newblock A Generative Model for Relation Extraction and Classification.
\newblock \emph{CoRR}, abs/2202.13229.

\bibitem[{Niu et~al.(2012)Niu, Zhang, R{\'{e}}, and
  Shavlik}]{DBLP:conf/vlds/NiuZRS12}
Niu, F.; Zhang, C.; R{\'{e}}, C.; and Shavlik, J.~W. 2012.
\newblock DeepDive: Web-scale Knowledge-base Construction using Statistical
  Learning and Inference.
\newblock In \emph{{VLDS}}, volume 884 of \emph{{CEUR} Workshop Proceedings},
  25--28. CEUR-WS.org.

\bibitem[{Raffel et~al.(2020)Raffel, Shazeer, Roberts, Lee, Narang, Matena,
  Zhou, Li, and Liu}]{DBLP:journals/jmlr/RaffelSRLNMZLL20}
Raffel, C.; Shazeer, N.; Roberts, A.; Lee, K.; Narang, S.; Matena, M.; Zhou,
  Y.; Li, W.; and Liu, P.~J. 2020.
\newblock Exploring the Limits of Transfer Learning with a Unified Text-to-Text
  Transformer.
\newblock \emph{J. Mach. Learn. Res.}, 21: 140:1--140:67.

\bibitem[{Rongali et~al.(2020)Rongali, Soldaini, Monti, and
  Hamza}]{DBLP:conf/www/RongaliSMH20}
Rongali, S.; Soldaini, L.; Monti, E.; and Hamza, W. 2020.
\newblock Don't Parse, Generate! {A} Sequence to Sequence Architecture for
  Task-Oriented Semantic Parsing.
\newblock In \emph{{WWW}}, 2962--2968. {ACM} / {IW3C2}.

\bibitem[{Rossiello et~al.(2021)Rossiello, Mihindukulasooriya, Abdelaziz,
  Bornea, Gliozzo, Naseem, and
  Kapanipathi}]{DBLP:conf/semweb/RossielloMABGNK21}
Rossiello, G.; Mihindukulasooriya, N.; Abdelaziz, I.; Bornea, M.~A.; Gliozzo,
  A.; Naseem, T.; and Kapanipathi, P. 2021.
\newblock Generative Relation Linking for Question Answering over Knowledge
  Bases.
\newblock In \emph{{ISWC}}, volume 12922 of \emph{Lecture Notes in Computer
  Science}, 321--337. Springer.

\bibitem[{Vrande\v{c}i\'{c} and Kr\"{o}tzsch(2014)}]{10.1145/2629489}
Vrande\v{c}i\'{c}, D.; and Kr\"{o}tzsch, M. 2014.
\newblock Wikidata: A Free Collaborative Knowledgebase.
\newblock \emph{Commun. ACM}, 57(10): 78–85.

\bibitem[{Wang et~al.(2021)Wang, Jiang, Bach, Wang, Huang, Huang, and
  Tu}]{DBLP:conf/acl/WangJBWHHT20}
Wang, X.; Jiang, Y.; Bach, N.; Wang, T.; Huang, Z.; Huang, F.; and Tu, K. 2021.
\newblock Improving Named Entity Recognition by External Context Retrieving and
  Cooperative Learning.
\newblock In \emph{{ACL/IJCNLP} {(1)}}, 1800--1812. Association for
  Computational Linguistics.

\bibitem[{Wu et~al.(2020)Wu, Petroni, Josifoski, Riedel, and
  Zettlemoyer}]{DBLP:conf/emnlp/WuPJRZ20}
Wu, L.; Petroni, F.; Josifoski, M.; Riedel, S.; and Zettlemoyer, L. 2020.
\newblock Scalable Zero-shot Entity Linking with Dense Entity Retrieval.
\newblock In \emph{{EMNLP} {(1)}}, 6397--6407. Association for Computational
  Linguistics.

\bibitem[{Wu, Zhang, and Li(2022)}]{DBLP:conf/acl/0001ZL22}
Wu, X.; Zhang, J.; and Li, H. 2022.
\newblock Text-to-Table: {A} New Way of Information Extraction.
\newblock In \emph{{ACL} {(1)}}, 2518--2533. Association for Computational
  Linguistics.

\bibitem[{Zhong and Chen(2021)}]{DBLP:conf/naacl/ZhongC21}
Zhong, Z.; and Chen, D. 2021.
\newblock A Frustratingly Easy Approach for Entity and Relation Extraction.
\newblock In \emph{{NAACL-HLT}}, 50--61. Association for Computational
  Linguistics.

\bibitem[{Zhou et~al.(2021)Zhou, Naseem, Astudillo, Lee, Florian, and
  Roukos}]{DBLP:conf/emnlp/ZhouNALFR21}
Zhou, J.; Naseem, T.; Astudillo, R.~F.; Lee, Y.; Florian, R.; and Roukos, S.
  2021.
\newblock Structure-aware Fine-tuning of Sequence-to-sequence Transformers for
  Transition-based {AMR} Parsing.
\newblock In \emph{{EMNLP} {(1)}}, 6279--6290. Association for Computational
  Linguistics.

\end{thebibliography}

\end{document}